\documentclass[sigconf]{acmart}
\NewCommandCopy\ThinkWiseEmail\email
\renewcommand{\email}[2][]{\ThinkWiseEmail[#1]{#2}\affiliation{\institution{}\city{}\country{}}}

\AtBeginDocument{}

\copyrightyear{2026}
\acmYear{2026}
\setcopyright{cc}
\setcctype{by}
\acmConference[MM '26]{Proceedings of the 34th ACM International Conference on Multimedia}{November 10--14, 2026}{Rio de Janeiro, Brazil}
\acmBooktitle{Proceedings of the 34th ACM International Conference on Multimedia (MM '26), November 10--14, 2026, Rio de Janeiro, Brazil}
\acmDOI{10.1145/3767308.3835837}
\acmISBN{979-8-4007-2213-4/2026/11}

\usepackage{algorithm}
\usepackage{tabularx}
\usepackage{algpseudocode} 
\usepackage[table]{xcolor}
\usepackage[normalem]{ulem}
\usepackage{booktabs}
\usepackage{enumitem}
\usepackage{multirow}
\usepackage{graphicx}
\usepackage[most]{tcolorbox}
\usepackage{float}
\setlist[itemize]{nosep}
\begin{document}

\title{ThinkWise: Learning When to Think in Large Multimodal Reasoning Models via Visual and Difficulty Perception}











\author{
Wenkang Zhou\textsuperscript{$\dagger$},
Chen Yang\textsuperscript{$\dagger$,1},
Jiaheng Peng\textsuperscript{1},
Xitong Li,
Hongzhi Yin\textsuperscript{2},
Yunke Zhang\textsuperscript{1},
Lu Yu\textsuperscript{3,$*$},
Peng-Fei Zhang\textsuperscript{4,$*$},
Xuhui Ren\textsuperscript{1,$*$},
Shijie Zhang\textsuperscript{$*$}
\\
\textsuperscript{$\dagger$}Equal Contribution.
\textsuperscript{$*$}Corresponding Author.
\\
\textsuperscript{1}Honor Device Co., Ltd \quad
\textsuperscript{2}The University of Queensland \quad
\textsuperscript{3}City University of Hong Kong \quad
\textsuperscript{4}Tongji University
\\
\url{https://github.com/Zhouwkk/ThinkWise}
}

\settopmatter{printacmref=false}
\renewcommand{\shortauthors}{Wenkang Zhou et al.}

\begin{abstract}
Large Multimodal Reasoning Models (LMRMs) extend large language models to heterogeneous modalities, demonstrating strong potential across a wide range of open-ended tasks. Reinforcement Learning with Verifiable Rewards (RLVR) has been widely adopted to improve reasoning performance in LMRMs; however, existing approaches overlook the critical role of visual perception uncertainty in guiding reasoning processes. We term this limitation Reasoning Allocation Misalignment, where the misalignment between visual perception uncertainty and reasoning depth leads to over-allocation of deep reasoning under uncertain visual grounding, resulting in unnecessarily long reasoning trajectories and degraded performance. To address this issue, we propose ThinkWise, a perception-aware RLVR framework that jointly models task difficulty and visual perception state to regulate reasoning depth. Specifically, we introduce an instance-adaptive visual perception modeling mechanism that enables context-aware calibration of perception uncertainty, leading to more accurate perception estimation. We further design a structured reward formulation that jointly considers perception, correctness, and task difficulty. In particular, an answer reward provides supervision for correctness, a perception- and difficulty-conditioned length reward adaptively regulates reasoning depth, and a perception reward encourages trajectories with stronger visual grounding. Extensive experiments on multiple multimodal reasoning benchmarks demonstrate that ThinkWise consistently improves the trade-off between accuracy and reasoning efficiency across both 3B and 7B models. These results highlight the importance of perception-aware reasoning and establish a principled framework for advancing LMRMs.
\end{abstract}



\maketitle

\section{Introduction}
\label{sec:intro}

Large Multimodal Reasoning Models (LMRMs) extend the generative capabilities of large language models to heterogeneous modalities, enabling unified understanding and reasoning over text, images, and other signals, and thus holding great promise for a wide range of open-ended tasks \cite{yao2025mmreason,li2025perception,huang2025vision,peng2025lmm,yang2025r1,chen2025r1v,shen2025vlm,meng2025mm,wang2025skywork,zhu2025active,zhang2024universal,9335490,zhang2026hierarchical,11644733}. At the core of these capabilities lies multimodal reasoning, which can be naturally decomposed into two fundamental components: multimodal perception and logical reasoning \cite{zhou2025proreason,xiao2025perception}. Prior studies \cite{chu2025qwenlookagainguiding,xu2025mitigating,jian-etal-2025-look} have shown that accurate visual perception is often the primary bottleneck, as it provides the essential visual basis upon which effective downstream reasoning critically depends.

Recent advances in Reinforcement Learning with Verifiable Rewards (RLVR) have demonstrated strong effectiveness in enhancing reasoning in large language models (LLMs), and have emerged as a dominant paradigm for improving multimodal reasoning capabilities of MLRMs by training them solely with rewards derived from the correctness of final answers \cite{chen2025r1v,shen2025vlm,shen2025vlm,peng2025lmm}. Given the critical role of visual perception, namely extracting task-relevant information from the image, more recent approaches have introduced dedicated mechanisms to strengthen this capability. One line of work \cite{xia2025visionaryr1mitigatingshortcutsvisual,zhu2025active} enforces structured visual outputs to anchor perceptual grounding during reasoning. Another line \cite{chu2025qwenlookagainguiding,jian-etal-2025-look,wan2025srpoenhancingmultimodalllm}  promotes visual re-attention to mitigate attention drift over long reasoning trajectories. The third line \cite{xiao2025perception,wang2025skywork}  uses external visual perception rewards to alleviate reward sparsity and mitigate hallucinations.

Despite extensive efforts showing that improving visual perception enhances reasoning performance, optimizing perception in isolation remains insufficient for effective multimodal reasoning. A critical limitation lies in the failure to effective alignment visual perception quality and reasoning depth, which we term \textbf{Reasoning Allocation Misalignment}, where reasoning depth is allocated without considering whether a confident perceptual foundation has been established for the given task. In conventional LLM settings, models tend to allocate longer reasoning trajectories to more challenging problems, implicitly trading increased reasoning depth for improved accuracy. However, this situation becomes more complex in multimodal scenarios, where uncertainty in visual perception can introduce additional complexity. When reasoning depth is governed solely by task difficulty, models may over-allocate deep reasoning to cases with uncertain visual perception, leading to unnecessarily long trajectories and degraded performance. Although prior work \cite{xiao2025fast} extends task difficulty to incorporate externally defined measures of both problem complexity and visual perception difficulty, the absence of intrinsic perceptual assessment limits its effectiveness, and thereby highlighting the need for a more principled, perception-aware reasoning paradigm.

Building upon the above observations, a central challenge in multimodal reasoning is how to jointly model visual perception uncertainty and task difficulty to guide reasoning depth allocation. Addressing this requires moving beyond difficulty-driven strategies and explicitly incorporating perceptual uncertainty into the reasoning process. Our key insight is simple yet fundamental: deeper reasoning should be encouraged only when additional computation is both necessary and supported by sufficient visual involvement, namely when the task is difficult and the current trajectory remains strongly grounded in visual perception. Concretely, we design multiple \textbf{perception–correctness constraints} to regulate reasoning behavior. When visual perception is strong, incorrect trajectories, especially on difficult problems, are encouraged to continue reasoning to refine their responses, while correct trajectories are encouraged to remain concise to avoid unnecessary computation. In contrast, when visual perception is weak, deeper reasoning is discouraged regardless of correctness, as extended reasoning under insufficient visual support tends to amplify unreliable perceptual premises or reinforce shortcut-like solutions that are not grounded in the image. This perception-aware reasoning paradigm improves reasoning effectiveness by mitigating ineffective deep reasoning and reducing deductions caused by weakly visual perception, while also enhancing efficiency by avoiding redundant computation when deeper reasoning is unlikely to be beneficial.

To this end, we propose \textit{ThinkWise}, a perception-aware RLVR framework for LMRMs. Our key innovation lies in moving beyond outcome-only supervision by explicitly incorporating trajectory-level perceptual signals into reasoning optimization, thereby enabling adaptive control over when deeper reasoning is beneficial. Specifically, we use Modality Attention Ratio (MAR) as a lightweight endogenous proxy, motivated by information-theoretic and internal-state perspectives \cite{Wu2026HowVB}, to estimate model's visual perception uncertainty. Building on this, we propose an instance-adaptive visual perception modeling mechanism that calibrates perception uncertainty in a context-aware manner, enabling more accurate perception signals. Finally, we propose a structured reward formulation that explicitly models perception–correctness constraints to guide reasoning behavior. The reward comprises three complementary components: an answer reward for correctness, a visual- and difficulty-perception length reward that adaptively regulates reasoning depth, and a perception reward that further promotes trajectories with stronger visual grounding. Through this design, ThinkWise elevates conventional outcome-only supervision into a structured, perception-aware training paradigm that enables principled reasoning allocation under perceptual reliability.
We summarize our contributions as follows:
\begin{itemize}
    \item We identify a fundamental limitation in multimodal reasoning, termed Reasoning Allocation Misalignment, where reasoning depth is allocated without considering visual perception, leading to suboptimal reasoning performance.
    \item We propose ThinkWise, a perception-aware RLVR framework that jointly models task difficulty and visual perception state to regulate reasoning depth, enabling adaptive reasoning allocation based on both necessity and perceptual reliability.
     \item We empirically evaluate ThinkWise on multiple MLLM reasoning benchmarks, demonstrating consistent improvements over strong RLVR baselines.
\end{itemize}

\section{Preliminary}
\label{sec:Preliminary}

\subsection{\textbf{Problem Formulation.}}
We study multimodal reasoning in the context ofLarge Multimodal Reasoning Models (LMRMs). Let
\(
\mathcal{D}=\{(v_i, q_i, a_i)\}_{i=1}^{N}
\)
denote a multimodal reasoning dataset, where \(v_i\), \(q_i\), and \(a_i\) represent the image, question, and reference answer of the \(i\)-th sample, respectively. Given an input pair \(x_i = (v_i,q_i)\), the policy model (i.e., LMRM) \(\pi_\theta\) generates a response sequence
\(
o_i
\) that aims to reach the ground-truth answer $a_i$. 
This reasoning process can be decomposed into two stages: (1) multimodal perception, which extracts and grounds relevant information from inputs $x_i$, and (2) textual reasoning, which operates on the perceived information to perform step-by-step inference and produce the final answer. Crucially, the effectiveness of reasoning depends on the reliability of visual perception, making the alignment between perception and reasoning a key factor in multimodal reasoning.

\subsection{\textbf{Reinforcement Learning with Verifiable Rewards (RLVR).}}
RLVR is a training paradigm that optimizes a LMRM using rewards derived from verifiable outcomes. Given an input $x_i$, a policy model $\pi_\theta$ generates a response trajectory $o_i \sim \pi_\theta(\cdot \mid x_i)$, and a reward function $R(x_i, o_i)$ evaluates the correctness of the final output based on a task-specific verifier. The training objective is to maximize the expected reward:
\[
\mathcal{L}_{\text{RLVR}}(\theta) = \mathbb{E}_{x_i \sim \mathcal{D},\, o_i \sim \pi_\theta(\cdot \mid x_i)} \left[ R(x_i, o_i) \right].
\]

\textit{\textbf{Group Relative Policy Optimization(GRPO).}}
GRPO~\cite{shao2024deepseekmath} is a widely used group-based policy optimization method in outcome-only RLVR.
Formally, for each input \(x_i\), we samples \(G\) candidate trajectories \(\{o_i^j\}_{j=1}^{G}\) from the old policy model \(\pi_{\theta_{\mathrm{old}}}\), and each trajectory is assigned a reward $R_i^j = R(x_i, o_i^j)$ . GRPO first computes group-relative advantages by normalizing rewards within the sampled group:
\begin{equation}
\hat{A}_i=
\frac{R_i^j-\mathrm{mean}(\mathbf{R})}{\mathrm{std}(\mathbf{R})},
\qquad
\mathbf{R}=\{R_i^1,\dots,R_i^G\},
\label{eq:group_adv}
\end{equation}

After obtaining the advantages, the policy model are then updated by maximizing the following objective:
\begin{equation}
\begin{aligned}
J_(\theta)
&= \mathbb{E}_{x_i \in \mathcal{D},\, \{o_i^j\}_{j=1}^{G}}
\Bigg[
    \frac{1}{G} \sum_{j=1}^{G} \frac{1}{|o_i^j|}
    \sum_{t=1}^{|o_i^j|}
    \Big(
        g_{j,t}(\theta)
        - \beta\, D_{\mathrm{KL}}(\pi_\theta \,\|\, \pi_{\mathrm{ref}})
    \Big)
\Bigg], \\
g_{j,t}(\theta)
&= \min\Big(
    r_{j,t}(\theta)\,\hat{A}_i,\,
    \operatorname{clip}(r_{j,t}(\theta), 1-\epsilon, 1+\epsilon)\,\hat{A}_i
\Big).
\end{aligned}
\label{eq:grpo_obj}
\end{equation}
where \(r_{i,t}(\theta)\) denotes the probability ratio for the \(i\)-th trajectory at time step \(t\), \(\pi_{\mathrm{ref}}\) is the is a fixed reference model typically initialized from the initial policy.

\begin{figure*}[t]
    \centering
    \includegraphics[width=\linewidth]{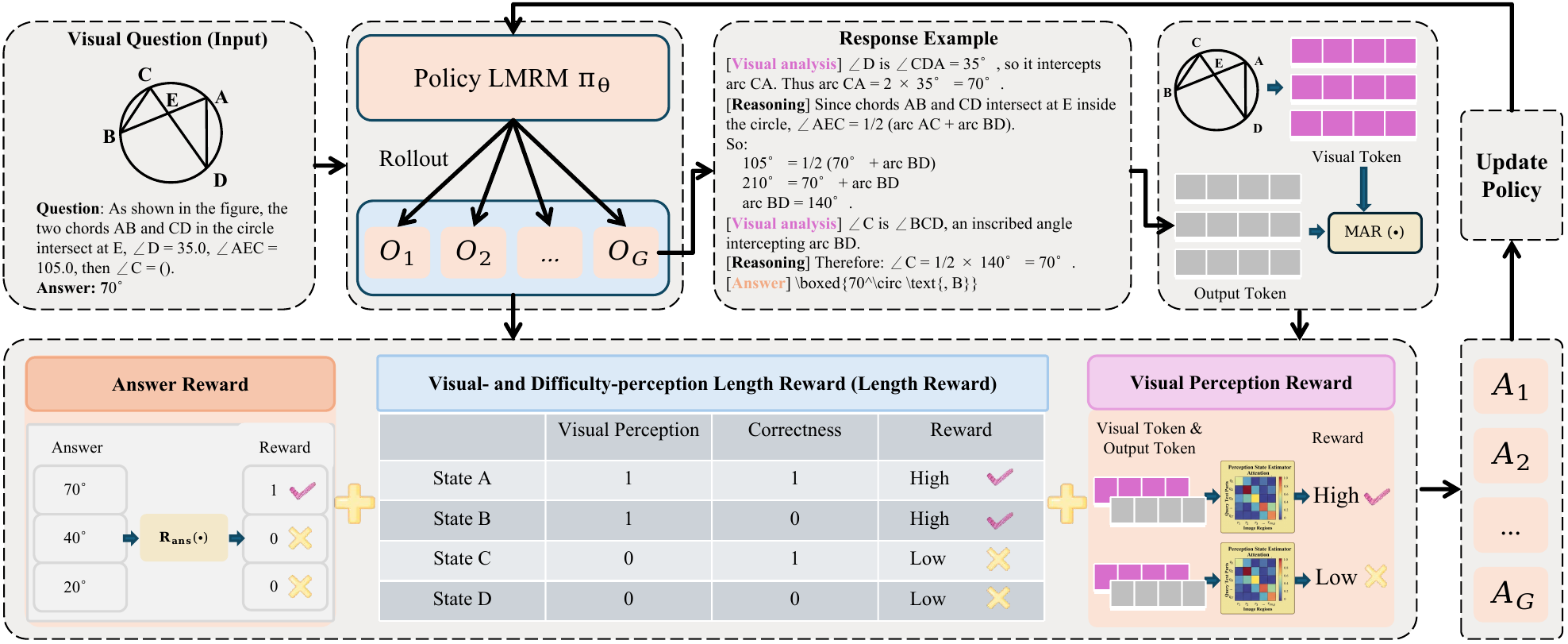}
    \caption{Overview of the ThinkWise training pipeline.}
    \label{fig:framework}
\end{figure*}
In standard GRPO, the reward $R_i$ for the $i$-th sampled output is typically defined solely by final-answer correctness, where $R_i \in \{0, 1\}$. While this sparse, outcome-driven supervision is highly effective for text-centric reasoning, relying strictly on it introduces a critical credit assignment problem in multimodal environments. In such settings, an incorrect response may originate from either flawed logical reasoning or inaccurate visual perception. Consequently, a binary outcome is insufficient to determine whether the model should be encouraged to allocate more computational resources (i.e., "think longer") to resolve the query. To address this limitation, ThinkWise preserves the core GRPO optimization objective while redesigning the sequence-level reward $R_i$ to jointly account for visual perception uncertainty and task difficulty.

\section{Methodology}
\label{sec:Methodology}

In this section, we present the technical details of ThinkWise, a perception-aware RLVR framework that addresses reasoning allocation misalignment. The key innovation of ThinkWise lies in modeling trajectory-level perceptual uncertainty, jointly considering visual perception and task difficulty, to adaptively regulate reasoning depth beyond conventional outcome-only supervision. We first introduce the Modality Attention Ratio (MAR) as a lightweight and scalable endogenous proxy to quantify the grounding quality. To effectively use MAR for reward design, we discretize it into a binary perception state via an instance-adaptive threshold derived from group-wise statistics. This design avoids reliance on fixed thresholds, improves robustness across tasks with varying degrees of visual dependence, and provides a reliable, context-aware signal for perception-aware reasoning. Building on this visual perception signal, we design a structured reward mechanism that explicitly captures perception–correctness conditions to regulate reasoning behavior. The reward consists of three components: an answer reward for correctness, a \emph{perception- and difficulty-conditioned length reward} that adaptively controls reasoning depth, and a visual perception reward that favors trajectories exhibiting strong visual perception. Together, these designs transform conventional outcome-only supervision into a structured, perception-aware training paradigm that aligns reasoning depth with both visual perception and task difficulty, enabling more effective multimodal reasoning.

\subsection{Estimating Visual Perception uncertainty via the Modality Attention Ratio}

To estimate the visual perception uncertainty of the current trajectory, we use the Modality Attention Ratio (MAR) as a proxy for visual engagement. Intuitively, stronger visual engagement corresponds to greater attention from output tokens to image tokens. 

Formally, consider a model with $L$ Transformer layers, and let $\mathcal{V}_{\text{img}}$ denote the set of image tokens. For the output token at position $k$ (i.e., token in $o_i$ ) in layer $l$, we define its aggregated attention score over all visual tokens in Eq.~\eqref{eq:visual_attention_score}:
\begin{equation}
s_k^{(l)}=\sum_{s\in\mathcal{V}_{\text{img}}}\mathbf{A}_{k,s}^{(l)},
\label{eq:visual_attention_score}
\end{equation}
where $\mathbf{A}_{k,s}^{(l)}$ denotes the attention weight from position $k$ to visual token $s$ in layer $l$. To mitigate noise from shallow feature alignment and interference induced by deep language-dominant reasoning, we use a middle-layer subset $\mathcal{L}_{\text{mid}}$ by default and compute the average MAR within a computation window $T$, which is defined as:
\begin{equation}
    \mathrm{MAR}(o)=\frac{1}{|\mathcal{L}_{\text{mid}}|}\sum_{l\in\mathcal{L}_{\text{mid}}}\frac{1}{T}\sum_{t=1}^{T}s_t^{(l)}.
\label{eq:mar_definition}
\end{equation}

Note that $MAR(o)$ characterizes the model's average level of visual engagement during the early stage of the current trajectory. 

\subsection{Instance-Adaptive Visual Perception Modelling}

To apply MAR as a control signal for reward allocation, we further discretize its continuous values into a binary visual perception state. A straightforward approach is to adopt a global threshold. However, multimodal tasks exhibit substantial variability in their intrinsic dependence on visual information. For example, geometry reasoning typically requires strong visual involvement, whereas certain commonsense questions may rely less on visual information. Consequently, a single global threshold fails to generalize across instances (i.e., input pair $x_i$) and may misclassify valid trajectories in low-visual-dependency tasks as weakly visually engaged. In light of this, we leverage the group-wise sampling mechanism and introduce an instance-adaptive visual-engagement criterion. For $G$ trajectories sampled from the same input $x_i$, we compute the group-wise mean and standard deviation of MAR, as defined: Eq.~\eqref{eq:group_mar_stats}:
\begin{equation}
    \mu_{\text{mar}}^{(i)}=\frac{1}{G}\sum_{j=1}^{G}\mathrm{MAR}(o_i^j), \quad
\sigma_{\text{mar}}^{(i)}=\sqrt{\frac{1}{G}\sum_{j=1}^{G}\left(\mathrm{MAR}(o_i^j)-\mu_{\text{mar}}^{(i)}\right)^2}.
\label{eq:group_mar_stats}
\end{equation}
We define an instance-adaptive visual engagement threshold as a lower bound computed from group-level statistics, formulated as:

\begin{equation}
\tau_{\text{mar}}^{(i)}=\mu_{\text{mar}}^{(i)}-\alpha\cdot\sigma_{\text{mar}}^{(i)}
\label{eq:mar_threshold},
\end{equation}
where $\alpha$ is a hyperparameter controlling the strictness of the threshold. The final binary state assignment is given by Eq.~\eqref{eq:mar_binary_state}:
\begin{equation}
    e(o_i^j)=\mathbf{1}\big[\mathrm{MAR}(o_i^j)>\tau_{\text{mar}}^{(i)}\big].
\label{eq:mar_binary_state}
\end{equation}

Importantly, the threshold is defined relative to the MAR distribution within each sampled group, and the resulting visual perception state is thus interpreted as a context-aware control signal rather than an absolute measure of perceptual correctness. This design enables instance-adaptive discrimination of trajectories based on their relative visual engagement, identifying those that provide sufficient perceptual grounding to support deeper reasoning. 

\subsection{Perception-Conditioned Reward Design}
\label{Perception-Conditioned Reward Design}
With the carefully designed estimation methods for visual perception of each trajectory, we introduce ThinkWise, an RLVR framework that introduces a perception-conditioned reward formulation to explicitly incorporate visual perception and difficulty into the learning signal for guiding reasoning allocation, as illustrated in Algorithm~\ref{alg:ThinkWise}. The overall reward consists of three components, each serving a distinct role in shaping reasoning behavior. Specifically, $R_{\text{ans}}$ provides the base supervision for answer correctness; $R_{\text{len}}$ adaptively modulates reasoning length conditioned on both visual perception uncertainty and task difficulty; and $R_{\text{perc}}$ further encourages trajectories with stronger visual engagement beyond correctness alone. Together, these components enable a principled integration of perception and reasoning signals. The total reward is defined in Eq.~\eqref{eq:total_reward}:

\begin{equation}
    R_{\text{total}}(o_i^j)=R_{\text{ans}}(o_i^j)+\gamma_{\text{len}}R_{\text{len}}(o_i^j)+\gamma_{\text{perc}}R_{\text{perc}}(o_i^j),
\label{eq:total_reward}
\end{equation}
where $\gamma_{\text{len}}$ and $\gamma_{\text{perc}}$ are hyperparameters that balance the three terms.

\subsubsection{Answer Reward}
The base correctness reward is computed by a rule-based verifier that checks whether the final generated answer matches the reference answer under the predefined equivalence rule. Formally, the answer reward is defined in Eq.~\eqref{eq:answer_reward}:
\begin{equation}
    R_{\text{ans}}(o_i^j)=
    \begin{cases}
    1, & \mathrm{eq}(a_i,o_i^j)=1,\\
    0, & \text{otherwise},
    \end{cases}
    \label{eq:answer_reward}
\end{equation}
where $a_i$ denotes the reference answer, and $\mathrm{eq}(a_i, o_i^j)$ is the answer equivalence function.

\subsubsection{Visual- and Difficulty-perception Length Reward (Length Reward)}

It is designed to regulate whether longer reasoning should be encouraged, rather than to uniformly favor either short or long responses. Its allocation depends jointly on three factors: the visual perception uncertainty of the current trajectory, answer correctness, and task difficulty. Intuitively, longer reasoning is encouraged only when the trajectory exhibits sufficient perceptual support yet still fails to produce the correct answer, especially on difficult problems. By contrast, when the answer is already correct or the visual perception is weak, additional length should be discouraged or left unrewarded. The four perception-aware constraints are summarized in Table~\ref{tab:length_reward_states}. Formally, the length reward is defined in Eq.~\eqref{eq:r_len}:

\begin{table}[t!]
\centering
\caption{Four constraints used in the visual- and difficulty-perception length reward. 
Here, $e(o)=1$ and $e(o)=0$ denote perception-strong and perception-weak trajectories.
}
\label{tab:length_reward_states}
\resizebox{\columnwidth}{!}{\begin{tabular}{c c c p{4.8cm}}
\toprule
State & Perception State & Correctness & Length-Reward Behavior \\
\midrule
A & $e(o)=1$ & $R_{\text{ans}}(o)=1$ & Favor concise reasoning and suppress unnecessary verbosity. \\
B & $e(o)=1$ & $R_{\text{ans}}(o)=0$ & Encourage longer reasoning; the incentive is stronger on more difficult problems. \\
C & $e(o)=0$ & $R_{\text{ans}}(o)=1$ & Discourage overly long reasoning, as this may reflect shortcut behavior. \\
D & $e(o)=0$ & $R_{\text{ans}}(o)=0$ & Assign no additional length-based incentive. \\
\bottomrule
\end{tabular}
}
\end{table}

\begin{equation}
R_{\mathrm{len}}(o_i^j)=
\begin{cases}
\operatorname{clip}(1-\rho,\,-1,\,1), 
& e(o_i^j)=1,\;R_{\text{ans}}(o_i^j)=1 \\
d_i \cdot \operatorname{clip}(\rho-1,\,0,\,1), 
& e(o_i^j)=1,\;R_{\text{ans}}(o_i^j)=0 \\
-\dfrac{1+d_i}{2}\cdot \operatorname{clip}(\rho-1,\,0,\,1), 
& e(o_i^j)=0,\;R_{\text{ans}}(o_i^j)=1 \\
0, 
& e(o_i^j)=0,\;R_{\text{ans}}(o_i^j)=0
\end{cases}
\label{eq:r_len}
\end{equation}
where $\rho=L(o_i^j)/L_{\mathrm{avg}}$ denotes the relative reasoning length, with $L_{\mathrm{avg}}$ being the group-wise average trajectory length, and $d_i = 1-\mathrm{PassRate}_i$ is a difficulty factor derived from the pass rate of problem $i$. As shown in Table~\ref{tab:length_reward_states}, $d_i$ strengthens the incentive for extended reasoning in State B and scales the penalty on overly long reasoning in State C according to problem difficulty. 

\subsubsection{Visual Perception Reward}

While the length reward regulates reasoning depth, it does not explicitly distinguish between correct trajectories supported by sufficient visual engagement and correct trajectories that may arise from shortcut-like behavior. To address this limitation, we introduce a separate perception reward $R_{\text{perc}}$, which is applied only to correct trajectories and further differentiates them according to their visual engagement.

Formally, the perception reward is defined in Eq.~\eqref{eq:r_perc}:
\begin{equation}
    R_{\text{perc}}(o_i^j)=
    \begin{cases}
    +\gamma_1\widetilde{\mathrm{MAR}}(o_i^j), & e(o_i^j)=1,\ R_{\text{ans}}(o_i^j)=1,\\
    -\gamma_2\widetilde{\tau}_{\text{mar}}, & e(o_i^j)=0,\ R_{\text{ans}}(o_i^j)=1,\\
    0, & R_{\text{ans}}(o_i^j)=0,
    \end{cases}
\label{eq:r_perc}
\end{equation}
where $\widetilde{\mathrm{MAR}}(o_i^j)$ and $\widetilde{\tau}_{\text{mar}}$ denote the group-normalized MAR score and threshold, respectively, as defined in Eq.~\eqref{eq:normalized_mar}.
\begin{equation}
    \widetilde{\mathrm{MAR}}(o_i^j)=\frac{\mathrm{MAR}(o_i^j)-\mathrm{MAR}_{\min}}{\mathrm{MAR}_{\max}-\mathrm{MAR}_{\min}},
    \quad
    \widetilde{\tau}_{\text{mar}}=\frac{\tau_{\text{mar}}-\mathrm{MAR}_{\min}}{\mathrm{MAR}_{\max}-\mathrm{MAR}_{\min}},
\label{eq:normalized_mar}
\end{equation}
where $\mathrm{MAR}_{\min}$ and $\mathrm{MAR}_{\max}$ are the minimum and maximum MAR values within the sampled trajectory group of the current input, and $\gamma_1$ and $\gamma_2$ control the strengths of the reward and penalty, respectively. For incorrect trajectories, $R_{\text{ans}}(o_i^j)=0$ already provides the main supervision signal, so no additional perception reward is applied. For correct trajectories, $R_{\text{perc}}$ distinguishes visually supported successes from perception-weak ones, discouraging shortcut-like solutions that bypass visual information.

\begin{algorithm}[t]
\caption{ThinkWise Training}
\label{alg:ThinkWise}
\begin{algorithmic}[1]
\Require Base model $\pi_\theta$, selected dataset $\mathcal{D}$, group size $G$, truncation window $T$
\State Initialize $\pi_{\mathrm{ref}} \gets \pi_\theta$
\For{$epoch = 1$ to $N$}
  \State Sample batch $\{(v_i,q_i,a_i)\} \sim \mathcal{D}$
  \State Generate grouped trajectories $\{o_i^j\}_{j=1}^{G} \sim \pi_\theta(v_i,q_i)$
    \State \textbf{Compute:}
    \State $\bullet$~$\mathrm{Acc}(o_i^j)=\mathrm{eq}(a_i,o_i^j)$
    \State $\bullet$~$\mathrm{MAR}(o_i^j)$ from multimodal attention over the first $T$ reasoning tokens
    \State $\bullet$~$\tau_{\mathrm{mar}}^{(i)}=\mu_{\mathrm{mar}}^{(i)}-\alpha\sigma_{\mathrm{mar}}^{(i)}$
    \State $\bullet$~$e(o_i^j)=\mathbf{1}[\mathrm{MAR}(o_i^j)>\tau_{\mathrm{mar}}^{(i)}]$
  \State \textbf{Compute rewards}:
  \[
  R_{\mathrm{total}}(o_i^j)=R_{\mathrm{ans}}(o_i^j)+\gamma_{\mathrm{len}}R_{\mathrm{len}}(o_i^j)+\gamma_{\mathrm{perc}}R_{\mathrm{perc}}(o_i^j)
  \]
 \State $\bullet$~$R_{\mathrm{len}}$: Eq.~\eqref{eq:r_len}
\State$\bullet$~$R_{\mathrm{perc}}$: Eq.~\eqref{eq:r_perc}
  \State Update policy:
  \[
  \max_{\theta}\ \mathbb{E}\!\left[\mathrm{GRPO}\!\left(\pi_\theta,\pi_{\mathrm{ref}},\hat{A},R_{\mathrm{total}}\right)\right]
  \]
\EndFor
\end{algorithmic}
\end{algorithm}

\section{Experiments}
In this section, we conduct experiments to verify the effectiveness of ThinkWise, aiming to answer the following research questions: 
\begin{itemize}
    \item \textbf{RQ1:} How does ThinkWise compare with existing multimodal reasoning post-training methods on standard benchmarks? 
    \item \textbf{RQ2:} What is the contribution of each key component in ThinkWise? 
    \item \textbf{RQ3:} Does ThinkWise generalize across different RL algorithms? 
    \item \textbf{RQ4:} How sensitive is ThinkWise to the main reward hyperparameters? 
    \item \textbf{RQ5:} How does ThinkWise affect the model's reasoning behavior in representative cases?
\end{itemize}

\subsection{Experimental Setup}
\begin{table*}[ht]
\centering
\caption{Main results on Qwen2.5-VL-3B and Qwen2.5-VL-7B. We report average accuracy (Acc.) and response length (Len.). 
}
\label{tab:main_results}
\resizebox{\textwidth}{!}{
\begin{tabular}{lccccccccccccccccc}
\toprule
\multirow{2}{*}{Method} & \multicolumn{2}{c}{MathVerse} & \multicolumn{2}{c}{MathVista} & \multicolumn{2}{c}{Counting} & \multicolumn{2}{c}{MMMU-Pro} & \multicolumn{2}{c}{MathVision} & \multicolumn{2}{c}{MMK12} & \multicolumn{2}{c}{We-Math} & \multicolumn{2}{c}{LogicVista} \\
\cmidrule(lr){2-3} \cmidrule(lr){4-5} \cmidrule(lr){6-7} \cmidrule(lr){8-9} \cmidrule(lr){10-11} \cmidrule(lr){12-13} \cmidrule(lr){14-15} \cmidrule(lr){16-17}
 & Acc.\,$\uparrow$ & Len.\,$\downarrow$ & Acc.\,$\uparrow$ & Len.\,$\downarrow$ & Acc.\,$\uparrow$ & Len.\,$\downarrow$ & Acc.\,$\uparrow$ & Len.\,$\downarrow$ & Acc.\,$\uparrow$ & Len.\,$\downarrow$ & Acc.\,$\uparrow$ & Len.\,$\downarrow$ & Acc.\,$\uparrow$ & Len.\,$\downarrow$ & Acc.\,$\uparrow$ & Len.\,$\downarrow$ \\
\midrule
\multicolumn{17}{c}{\textbf{Base Qwen2.5-VL-3B-Instruct}} \\
\midrule
Qwen2.5-VL-3B-Instruct & 16.24 & 229.2 & 29.95 & 162.2 & 20.50 & 201.9 & 9.61 & 267.1 & 5.13 & 291.3 & 24.38 & 421.3 & 14.99 & 220.5 & 18.90 & 188.6 \\
Curr-ReFT & 38.11 & 98.3 & 52.47 & 88.5 & 43.91 & 82.6 & 18.91 & 106.1 & 8.58 & 125.3 & 40.52 & 209.1 & 31.06 & 105.0 & 34.65 & 103.2 \\
RLCS & 28.03 & 712.0 & \uline{61.74} & 437.5 & 70.25 & 250.6 & 26.65 & 759.3 & 12.20 & 845.7 & 42.70 & 968.4 & 41.96 & 626.8 & 39.40 & 534.6 \\
VLAA & 39.79 & 718.1 & 51.19 & 332.2 & 55.25 & 185.8 & 21.99 & 381.6 & 9.13 & 877.8 & 40.21 & 673.2 & 43.02 & 660.3 & 32.47 & 388.6 \\
FAST & 39.93 & 287.3 & \textbf{67.21} & 146.3 & \uline{72.50} & 82.5 & 22.49 & 274.8 & \textbf{14.04} & 309.0 & 47.38 & 339.2 & 38.53 & 216.8 & 36.05 & 157.3 \\
MMR1 & 32.10 & 1078.9 & 56.40 & 644.3 & 47.06 & 315.2 & \uline{27.17} & 1365.5 & 10.81 & 1347.8 & 39.01 & 1327.2 & 51.23 & 859.6 & 35.12 & 883.9 \\
PAPO & \textbf{56.99} & 353.9 & 58.35 & 184.4 & 57.69 & 101.5 & 26.57 & 401.2 & 11.50 & 420.9 & \uline{57.12} & 529.6 & \uline{59.35} & 348.3 & \textbf{41.64} & 242.3 \\
\rowcolor{gray!10}
ThinkWise-3B & \uline{56.82} & 262.6 & 59.16 & 153.2 & \textbf{76.56} & 109.0 & \textbf{29.08} & 215.2 & \uline{12.32} & 327.1 & \textbf{67.15} & 342.6 & \textbf{59.73} & 258.5 & \uline{41.11} & 167.7 \\
\midrule
\multicolumn{17}{c}{\textbf{Base Qwen2.5-VL-7B-Instruct}} \\
\midrule
Qwen2.5-VL-7B-Instruct & 27.10 & 370.5 & 61.11 & 226.3 & 52.94 & 95.2 & 16.29 & 397.2 & 10.59 & 479.5 & 36.31 & 516.1 & 29.85 & 333.9 & 38.53 & 308.4 \\
Curr-ReFT & 37.60 & 147.8 & 63.48 & 129.0 & 62.25 & 97.7 & 19.35 & 182.4 & 10.93 & 202.3 & 47.57 & 241.3 & 44.70 & 134.6 & 42.25 & 160.5 \\
RLCS & 3.64 & 705.2 & \textbf{70.00} & 432.8 & 80.25 & 244.9 & 10.46 & 783.2 & 14.05 & 817.2 & 34.74 & 863.2 & 6.66 & 606.0 & 21.42 & 538.7 \\
VLAA & 43.93 & 329.0 & 63.49 & 216.9 & 75.75 & 95.3 & 24.43 & 367.8 & 12.41 & 441.4 & 44.67 & 492.1 & 47.79 & 309.3 & 43.26 & 286.8 \\
FAST & \uline{66.78} & 146.2 & 64.41 & 115.2 & 78.56 & 72.7 & 34.80 & 231.0 & 13.74 & 284.7 & \uline{76.33} & 174.2 & 66.08 & 130.0 & \textbf{46.36} & 162.7 \\
MMR1 & 33.45 & 1058.0 & 59.36 & 627.4 & 72.50 & 257.2 & 16.68 & 1312.6 & 13.91 & 1391.9 & 24.32 & 1236.7 & 44.18 & 829.0 & 31.49 & 889.8 \\
R1-Onevision & 17.63 & 273.4 & 45.17 & 197.7 & 84.44 & 95.6 & 16.63 & 235.9 & 8.63 & 289.7 & 25.59 & 320.4 & 27.05 & 220.1 & 24.58 & 210.8 \\
ReVisual-R1 & 55.35 & 2474.9 & 61.44 & 1529.1 & 67.19 & 890.2 & 31.66 & 2461.1 & \textbf{16.93} & 3369.9 & 37.36 & 3035.1 & 61.82 & 2135.1 & 40.10 & 2585.5 \\
PAPO & \textbf{67.06} & 378.6 & 66.79 & 213.0 & \uline{87.25} & 94.6 & \uline{34.81} & 398.4 & 13.80 & 465.9 & 72.32 & 441.4 & \uline{66.19} & 327.3 & 44.74 & 284.7 \\
\rowcolor{gray!10}
ThinkWise-7B & 66.07 & 344.7 &\uline{ 67.45} & 240.8 & \textbf{92.69} & 122.7 & \textbf{36.42} & 348.1 & \uline{14.84} & 410.7 & \textbf{79.67} & 428.3 & \textbf{67.92} & 306.1 & \uline{44.94} & 275.9 \\
\bottomrule
\end{tabular}
}
\end{table*}

\subsubsection{Training Data and Protocol.}
All models are trained on ViRL39\\~\cite{shen2025vlm} for one epoch with a learning rate of \(1\times10^{-6}\). The dataset contains 38K verifiable multimodal reasoning samples covering visual mathematics, STEM reasoning, and chart understanding. To improve training stability, we further adopt a curriculum-style sampling strategy. In the early stage of training, we filter out overly easy samples so that the model focuses on sufficiently challenging examples and gradually improves its reasoning capability. In the later stage, we filter out overly difficult samples so that training concentrates on learnable cases, helping the model better learn when to reason deeply and when to respond concisely. 

\subsubsection{Baselines.}
We use Qwen2.5-VL-3B-Instruct and Qwen2.5-VL-7B-Instruct as the backbone models. We compare ThinkWise against three groups of baselines: the original base models, existing multimodal reasoning post-training methods, and generic RL algorithms (GRPO \cite{shao2024deepseekmath}, Reinforce++ \cite{hu2025reinforcestabilizingcriticfreepolicy}, and PPO \cite{schulman2017proximal}) together with their ThinkWise-based variants. The compared post-training methods include RLCS~\cite{wei2025advancingmultimodalreasoningreinforcement}, VLAA~\cite{chen2025sftrlearlyinvestigation}, FAST~\cite{xiao2025fast}, MMR1~\cite{leng2025mmr1}, PAPO~\cite{wang2025perception}, Curr-ReFT~\cite{deng2025boosting}, R1-OneVision~\cite{yang2025r1}, and ReVisual-R1~\cite{chen2025advancing}. Since only 7B checkpoints are publicly available for R1-OneVision and ReVisual-R1, we include them only in the 7B setting.

\subsubsection{Benchmarks and Metrics.}
We evaluate ThinkWise on benchmarks spanning mathematical and geometric reasoning, multidisciplinary multimodal reasoning, logical reasoning, and counting. MathVista~\cite{lu2023mathvista}, MathVerse~\cite{zhang2024mathverse}, MathVision~\cite{wang2024measuringmultimodalmathematicalreasoning}, WeMath~\cite{qiao2024wemathdoeslargemultimodal}, and MM-K12~\cite{meng2025mm} measure broad multimodal reasoning ability, while LogicVista~\cite{xiao2024logicvista}, Counting\cite{li2023superclevrvirtualbenchmarkdiagnose}, and MMMU-Pro~\cite{yue2025mmmu} are used to assess visual grounding sensitivity. We report accuracy (Acc.) and average response length (Len.) for all benchmarks. All results are averaged over eight independent runs to account for output stochasticity.  

\subsection{Main Results (RQ1)}

\begin{figure}[t]
    \centering
    \includegraphics[width=\columnwidth]{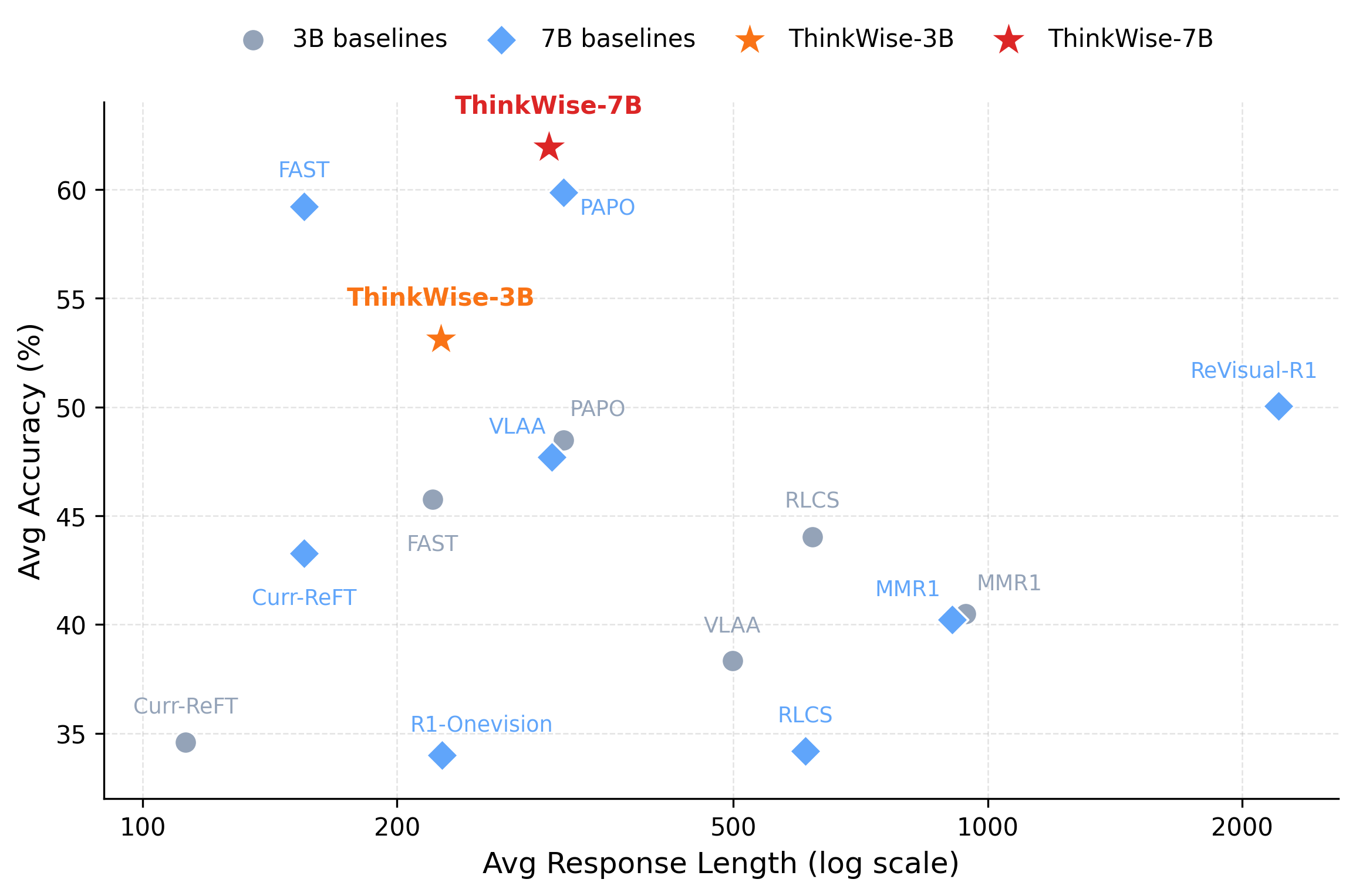}
    \caption{Accuracy versus average response length.
}
    \label{fig:efficiency_accuracy}
\end{figure}

In Table~\ref{tab:main_results}, ThinkWise achieves the highest average accuracy across all eight benchmarks at both the 3B and 7B scales. Compared with the strongest baseline, PAPO, it improves the average accuracy by 4.09\% and 2.13\%, respectively, while producing responses that are 16.8\% shorter on average across the two scales. Figure~\ref{fig:efficiency_accuracy} further illustrates the accuracy--efficiency trade-off across all methods, showing that ThinkWise occupies the Pareto-optimal region at both scales.

\subsubsection{Performance on perception-critical benchmarks.}
The largest improvements are observed on benchmarks that require reliable visual grounding for correct reasoning, especially Counting and MMK12, where ThinkWise achieves significant gains at both scales. These results are consistent with the hypothesis that explicitly coupling perception and reasoning leads to more effective use of visual information than outcome-only supervision.

\subsubsection{Comparison with diffuculty-aware method.}
Compared with FAST, which considers both task and visual difficulty, ThinkWise performs better on tasks that require sustained visual engagement (e.g., MMK12 and We-Math), and also outperforms FAST on MathVerse at the 3B scale. FAST remains stronger on MathVista, where symbolic or spatial reasoning appears to more important than visual grounding. This pattern is consistent with the design of ThinkWise, whose perception-aware reward is most effective when visual grounding remains a bottleneck throughout the reasoning process.

\subsubsection{Long-chain methods and ineffective deep reasoning.}
Long-chain methods such as MMR1 and ReVisual-R1 generate responses that are 3 to 8 times longer than those of ThinkWise, yet consistently obtain lower average accuracy. Without explicit alignment between perception and reasoning, longer reasoning chains may reinforce unreliable perceptual premises rather than improve the final answer. 

\begin{table*}[t]
\centering
\setlength{\tabcolsep}{3.5pt}
\caption{Ablation results on Qwen2.5-VL-3B and Qwen2.5-VL-7B. 
}
\label{tab:ablation_results}
\resizebox{\textwidth}{!}{
\begin{tabular}{lcccccccccccccccc}
\toprule
\multirow{2}{*}{Ablation}
& \multicolumn{2}{c}{MathVerse}
& \multicolumn{2}{c}{MathVista}
& \multicolumn{2}{c}{Counting}
& \multicolumn{2}{c}{MMMU-Pro}
& \multicolumn{2}{c}{MathVision}
& \multicolumn{2}{c}{MMK12}
& \multicolumn{2}{c}{We-Math}
& \multicolumn{2}{c}{LogicVista} \\
\cmidrule(lr){2-3}
\cmidrule(lr){4-5}
\cmidrule(lr){6-7}
\cmidrule(lr){8-9}
\cmidrule(lr){10-11}
\cmidrule(lr){12-13}
\cmidrule(lr){14-15}
\cmidrule(lr){16-17}
& Acc.\,$\uparrow$ & Len.\,$\downarrow$
& Acc.\,$\uparrow$ & Len.\,$\downarrow$
& Acc.\,$\uparrow$ & Len.\,$\downarrow$
& Acc.\,$\uparrow$ & Len.\,$\downarrow$
& Acc.\,$\uparrow$ & Len.\,$\downarrow$
& Acc.\,$\uparrow$ & Len.\,$\downarrow$
& Acc.\,$\uparrow$ & Len.\,$\downarrow$
& Acc.\,$\uparrow$ & Len.\,$\downarrow$ \\
\midrule
\multicolumn{17}{c}{\textbf{3B Ablation}} \\
\midrule
Qwen2.5-VL-3B-Instruct
& 16.24 & 229.2
& 29.95 & 162.2
& 20.50 & 201.9
& 9.61 & 267.1
& 5.13 & 291.3
& 24.38 & 421.3
& 14.99 & 220.5
& 18.90 & 188.6 \\

\rowcolor{gray!10}
ThinkWise-3B
& \textbf{56.82} & 262.6
& \uline{59.16} & 153.2
& \textbf{76.56} & 109.0
& \textbf{29.08} & 215.2
& \textbf{12.32} & 327.1
& \textbf{67.15} & 342.6
& \textbf{59.73} & 258.5
& \textbf{41.11} & 167.7 \\

w/o Length Reward
& \uline{56.73} & 319.9
& \textbf{61.34} & 205.6
& \uline{73.25} & 130.2
& \uline{27.25} & 334.5
& \uline{11.94} & 372.1
& \uline{66.95} & 476
& \uline{58.87} & 323.5
& \uline{38.79} & 255.1 \\

w/o Perception Reward
& 55.71 & 352.6
& 56.1 & 244.4
& 58.88 & 177.7
& 25.06 & 331.3
& 10.84 & 412.6
& 63.12 & 539.9
& 59.17 & 349
& 38.34 & 292.7 \\

w/o Curriculum
& 55.42 & 650.6
& 55.31 & 592.8
& 56.69 & 187.1
& 23.74 & 645.9
& 11.45 & 2063.7
& 60.48 & 620.4
& 58.29 & 696.2
& 36.91 & 425.6 \\

Vanilla GRPO
& 54.17 & 350.1
& 57.44 & 220.2
& 56.12 & 171.4
& 25.61 & 333.4
& 10.72 & 437.9
& 65.88 & 526.6
& 57.21 & 379.1
& 38.06 & 281.9 \\

\midrule
\multicolumn{17}{c}{\textbf{7B Ablation}} \\
\midrule
Qwen2.5-VL-7B-Instruct
& 27.10 & 370.5
& 61.11 & 226.3
& 52.94 & 95.2
& 16.29 & 397.2
& 10.59 & 479.5
& 36.31 & 516.1
& 29.85 & 333.9
& 38.53 & 308.4 \\

\rowcolor{gray!10}
ThinkWise-7B
& \textbf{66.07} & 344.7
& \textbf{67.45} & 240.8
& \textbf{92.69} & 122.7
& \textbf{36.42} & 348.1
& \textbf{14.84} & 410.7
& \uline{79.67} & 428.3
& \textbf{67.92} & 306.1
& \uline{44.94} & 275.9 \\

w/o Length Reward
& \uline{65.79} & 407.8
& \uline{65.79} & 390.9
& \uline{88.19} & 92.3
& \uline{36.05} & 407.2
& \uline{14.07} & 464.3
& \textbf{80.74} & 478.8
& \uline{67.90} & 373.4
& 44.27 & 311.9 \\

w/o Perception Reward
& 62.90 & 322.4
& 64.76 & 259.6
& 84.44 & 101.1
& 35.54 & 320.0
& 13.94 & 375.4
& 73.81 & 528.0
& 66.11 & 295.5
& 44.85 & 339.2 \\

w/o Curriculum
& 63.72 & 500.3
& 64.05 & 454.5
& 80.81 & 199.8
& 35.60 & 672.3
& 12.56 & 2822.8
& 74.39 & 415.5
& 65.26 & 382.4
& \textbf{45.86} & 633.9 \\

Vanilla GRPO
& 63.96 & 384.3
& 62.10 & 278.7
& 84.50 & 100.1
& 35.30 & 460.5
& 12.94 & 515.1
& 74.28 & 534.2
& 67.70 & 361.2
& 43.43 & 345.5 \\
\bottomrule
\end{tabular}
}
\end{table*}

\subsubsection{Scaling behavior.}
The gains from 3B to 7B are most pronounced on Counting and MMK12, suggesting that tasks requiring broader visual understanding and stronger knowledge integration benefit more from increased model capacity. Across both scales, ThinkWise ranks first or second on nearly all benchmarks, indicating that the perception-aware design generalizes well across model sizes.

\subsection{Ablation Study (RQ2)}

Table~\ref{tab:ablation_results} summarizes the ablation results of the three core components of ThinkWise on the 3B and 7B backbones.
\begin{table*}[ht]
\centering
\setlength{\tabcolsep}{3.5pt}
\caption{Cross-algorithm generalization results on 3B models. 
}
\label{tab:cross_rl_3b}
\resizebox{\textwidth}{!}{
\begin{tabular}{llcccccccccccccccccc}
\toprule
\multirow{2}{*}{Optimizer} & \multirow{2}{*}{Variant}
& \multicolumn{2}{c}{MathVerse}
& \multicolumn{2}{c}{MathVista}
& \multicolumn{2}{c}{Counting}
& \multicolumn{2}{c}{MMMU-Pro}
& \multicolumn{2}{c}{MathVision}
& \multicolumn{2}{c}{MMK12}
& \multicolumn{2}{c}{We-Math}
& \multicolumn{2}{c}{LogicVista}
& \multirow{2}{*}{Avg. Acc.}
& \multirow{2}{*}{Avg. Len.} \\
\cmidrule(lr){3-4}
\cmidrule(lr){5-6}
\cmidrule(lr){7-8}
\cmidrule(lr){9-10}
\cmidrule(lr){11-12}
\cmidrule(lr){13-14}
\cmidrule(lr){15-16}
\cmidrule(lr){17-18}
&
& Acc.\,$\uparrow$ & Len.\,$\downarrow$
& Acc.\,$\uparrow$ & Len.\,$\downarrow$
& Acc.\,$\uparrow$ & Len.\,$\downarrow$
& Acc.\,$\uparrow$ & Len.\,$\downarrow$
& Acc.\,$\uparrow$ & Len.\,$\downarrow$
& Acc.\,$\uparrow$ & Len.\,$\downarrow$
& Acc.\,$\uparrow$ & Len.\,$\downarrow$
& Acc.\,$\uparrow$ & Len.\,$\downarrow$
& & \\
\midrule
\multirow{2}{*}{GRPO}
& Vanilla
& 54.17 & 350.1
& 57.44 & 220.2
& 56.12 & 171.4
& 25.61 & 333.4
& 10.72 & 437.9
& 65.88 & 526.6
& 57.21 & 379.1
& 38.06 & 281.9
& 45.65 & 337.58 \\
& \cellcolor{gray!10} ThinkWise
& \cellcolor{gray!10} 56.82 & \cellcolor{gray!10} 262.6
& \cellcolor{gray!10} 59.16 & \cellcolor{gray!10} 153.2
& \cellcolor{gray!10} 76.56 & \cellcolor{gray!10} 109.0
& \cellcolor{gray!10} 29.08 & \cellcolor{gray!10} 215.2
& \cellcolor{gray!10} 12.32 & \cellcolor{gray!10} 327.1
& \cellcolor{gray!10} 67.15 & \cellcolor{gray!10} 342.6
& \cellcolor{gray!10} 59.73 & \cellcolor{gray!10} 258.5
& \cellcolor{gray!10} 41.11 & \cellcolor{gray!10} 167.7
& \cellcolor{gray!10} 50.24 (+4.59) & \cellcolor{gray!10} 229.49 (-32.0\%) \\
\midrule
\multirow{2}{*}{Reinforce++}
& Vanilla
& 46.79 & 317.2
& 43.10 & 211.5
& 41.50 & 117.0
& 19.13 & 306.3
& 4.70 & 93.9
& 50.55 & 377.6
& 52.01 & 291.7
& 35.57 & 246.6
& 36.67 & 245.23 \\
& \cellcolor{gray!10} ThinkWise
& \cellcolor{gray!10} 57.72 & \cellcolor{gray!10} 304.0
& \cellcolor{gray!10} 60.16 & \cellcolor{gray!10} 188.9
& \cellcolor{gray!10} 76.00 & \cellcolor{gray!10} 142.2
& \cellcolor{gray!10} 26.74 & \cellcolor{gray!10} 305.5
& \cellcolor{gray!10} 13.82 & \cellcolor{gray!10} 349.3
& \cellcolor{gray!10} 70.41 & \cellcolor{gray!10} 361.1
& \cellcolor{gray!10} 60.14 & \cellcolor{gray!10} 284.5
& \cellcolor{gray!10} 41.30 & \cellcolor{gray!10} 236.8
& \cellcolor{gray!10} 50.79 (+14.12) & \cellcolor{gray!10} 271.54 (+10.7\%) \\
\midrule
\multirow{2}{*}{PPO}
& Vanilla
& 36.88 & 82.3
& 51.6 & 50.4
& 61 & 5.2
& 20.77 & 92.3
& 7.41 & 96.3
& 42.13 & 66.7
& 43.1 & 81.3
& 31.82 & 85.6
& 36.84 & 70.01 \\
& \cellcolor{gray!10} ThinkWise
& \cellcolor{gray!10} 50.93 & \cellcolor{gray!10} 79.8
& \cellcolor{gray!10} 55.27 & \cellcolor{gray!10} 123.6
& \cellcolor{gray!10} 46.94 & \cellcolor{gray!10} 45
& \cellcolor{gray!10} 24.02 & \cellcolor{gray!10} 75.7
& \cellcolor{gray!10} 7.96 & \cellcolor{gray!10} 158.1
& \cellcolor{gray!10} 58.01 & \cellcolor{gray!10} 63
& \cellcolor{gray!10} 54.36 & \cellcolor{gray!10} 107.1
& \cellcolor{gray!10} 35.35 & \cellcolor{gray!10} 97.1
& \cellcolor{gray!10} 41.61 (+4.77) & \cellcolor{gray!10} 93.68 (+33.8\%) \\
\bottomrule

\end{tabular}
}
\end{table*}

\subsubsection{Distinct roles of $R_{\mathrm{perc}}$ and $R_{\mathrm{len}}$.}
Removing $R_{\mathrm{perc}}$ consistently reduces accuracy on both backbones, confirming that perception-aware reward is critical for LMRMs. Its effect on response length is scale-dependent: on 3B, removing $R_{\mathrm{perc}}$ leads to longer responses, suggesting that the smaller model is more prone to drifting into longer but weakly grounded reasoning traces without explicit groun\\ding-aware preference; on 7B, the length change is smaller and less uniform, indicating that the larger model already has stronger intrinsic grounding ability, so the main degradation appears in answer quality rather than a consistent shift in length.

Removing $R_{\mathrm{len}}$ increases response length on both scales, confirming that it is the primary mechanism for suppressing unnecessarily long outputs. The associated accuracy drop is generally smaller than that caused by removing $R_{\mathrm{perc}}$, suggesting that $R_{\mathrm{len}}$ mainly regulates reasoning efficiency, whereas $R_{\mathrm{perc}}$ contributes more directly to reasoning quality. The stronger length increase on 3B further indicates that smaller models are more sensitive to length reward.

\subsubsection{Effect of the curriculum.}
Removing the curriculum yields the most unstable behavior on both backbones, with especially severe length explosion on visually demanding tasks. In these cases, longer reasoning does not improve performance, but instead amplifies errors from unreliable perceptual premises. 

\subsubsection{Comparison with vanilla GRPO.}
Compared with vanilla GRPO, ThinkWise consistently improves accuracy on both model scales, with particularly clear gains on visually demanding benchmarks such as Counting, MMMU-Pro, and MMK12. 

\subsection{Generalization Across RL Algorithms (RQ3)}
Table~\ref{tab:cross_rl_3b} evaluates the generalizability of ThinkWise across different RL algorithms on the 3B backbone. We consider three representative optimizers, namely GRPO, Reinforce++, and PPO. Across all settings, ThinkWise consistently outperforms the corresponding vanilla training, indicating that its effectiveness is not tied to any specific RL algorithm. Notably, the improvements are more pronounced on benchmarks that heavily rely on visual grounding (e,g., Counting, MMK12, MathVision, and MMMU-Pro), while relatively smaller gains are observed on tasks where visual perception plays a less critical role. Although accuracy improvements remain consistent across optimizers, the behavior of response length varies, suggesting that the accuracy benefits of ThinkWise are robust, whereas its impact on reasoning length is more sensitive to the choice of optimizer.

\subsection{Sensitivity Analysis (RQ4)}
\label{sec:sensitivity}
\begin{table*}[t]
\centering
\setlength{\tabcolsep}{3.5pt}
\caption{Sensitivity analysis of reward weights on 7B models. 
}
\label{tab:reward_weight_sensitivity}
\resizebox{\textwidth}{!}{
\begin{tabular}{lcccccccccccccccccc}
\toprule
\multirow{2}{*}{Setting}
& \multicolumn{2}{c}{MathVerse}
& \multicolumn{2}{c}{MathVista}
& \multicolumn{2}{c}{Counting}
& \multicolumn{2}{c}{MMMU-Pro}
& \multicolumn{2}{c}{MathVision}
& \multicolumn{2}{c}{MMK12}
& \multicolumn{2}{c}{We-Math}
& \multicolumn{2}{c}{LogicVista}
& \multirow{2}{*}{Avg. Acc.\,$\uparrow$}
& \multirow{2}{*}{Avg. Len.\,$\downarrow$} \\
\cmidrule(lr){2-3}
\cmidrule(lr){4-5}
\cmidrule(lr){6-7}
\cmidrule(lr){8-9}
\cmidrule(lr){10-11}
\cmidrule(lr){12-13}
\cmidrule(lr){14-15}
\cmidrule(lr){16-17}
& Acc.\,$\uparrow$ & Len.\,$\downarrow$
& Acc.\,$\uparrow$ & Len.\,$\downarrow$
& Acc.\,$\uparrow$ & Len.\,$\downarrow$
& Acc.\,$\uparrow$ & Len.\,$\downarrow$
& Acc.\,$\uparrow$ & Len.\,$\downarrow$
& Acc.\,$\uparrow$ & Len.\,$\downarrow$
& Acc.\,$\uparrow$ & Len.\,$\downarrow$
& Acc.\,$\uparrow$ & Len.\,$\downarrow$
& & \\
\midrule
S0 \ ($\gamma_{\mathrm{len}}=0.1,\ \gamma_{\mathrm{perc}}=0.1$)
& \textbf{66.07} & 344.7
& 67.45 & 240.8
& \uline{92.69} & 122.7
& \textbf{36.42} & 348.1
& \textbf{14.84} & 410.7
& \textbf{79.67} & 428.3
& \textbf{67.92} & 306.1
& \uline{44.94} & 275.9
& \textbf{58.751} & \uline{309.66} \\

S1 \ ($\gamma_{\mathrm{len}}=0.2,\ \gamma_{\mathrm{perc}}=0.1$)
& 62.31 & 308.2
& \textbf{68.25} & 199.1
& 91.00 & 97.5
& 33.50 & 255.9
& 12.27 & 348.8
& 75.65 & 329.4
& 65.32 & 265.0
& 43.32 & 222.1
& 56.453 & \textbf{253.25} \\

S2 \ ($\gamma_{\mathrm{len}}=0.1,\ \gamma_{\mathrm{perc}}=0.2$)
& \uline{65.34} & 353.6
& \uline{68.09} & 246.1
& \textbf{93.00} & 115.8
& \uline{36.06} & 365.8
& \uline{14.18} & 410.6
& \uline{79.60} & 429.9
& \uline{66.58} & 304.9
& \textbf{45.08} & 290.0
& \uline{58.491} & 314.59 \\
\bottomrule
\end{tabular}
}
\end{table*}

Table~\ref{tab:reward_weight_sensitivity} reports the results of three reward-weight settings on the 7B backbone, including the default setting S0 with $\gamma_{\mathrm{len}}=0.1$ and $\gamma_{\mathrm{perc}}=0.1$, a stronger length reward setting S1 with $\gamma_{\mathrm{len}}=0.2$, and a stronger perception reward setting S2 with $\gamma_{\mathrm{perc}}=0.2$. The results, shown in Table~\ref{tab:reward_weight_sensitivity}, indicate that S0 achieves the best trade-off between accuracy and response length among the three settings. The maximum average accuracy difference across all configurations is 2.30\%, which suggests that ThinkWise remains reasonably stable under moderate changes in reward weights. We therefore use S0 as the default setting in all remaining experiments.

\subsection{Case Study (RQ5)}
\begin{figure}[t]
    \centering
    \includegraphics[width=\columnwidth]{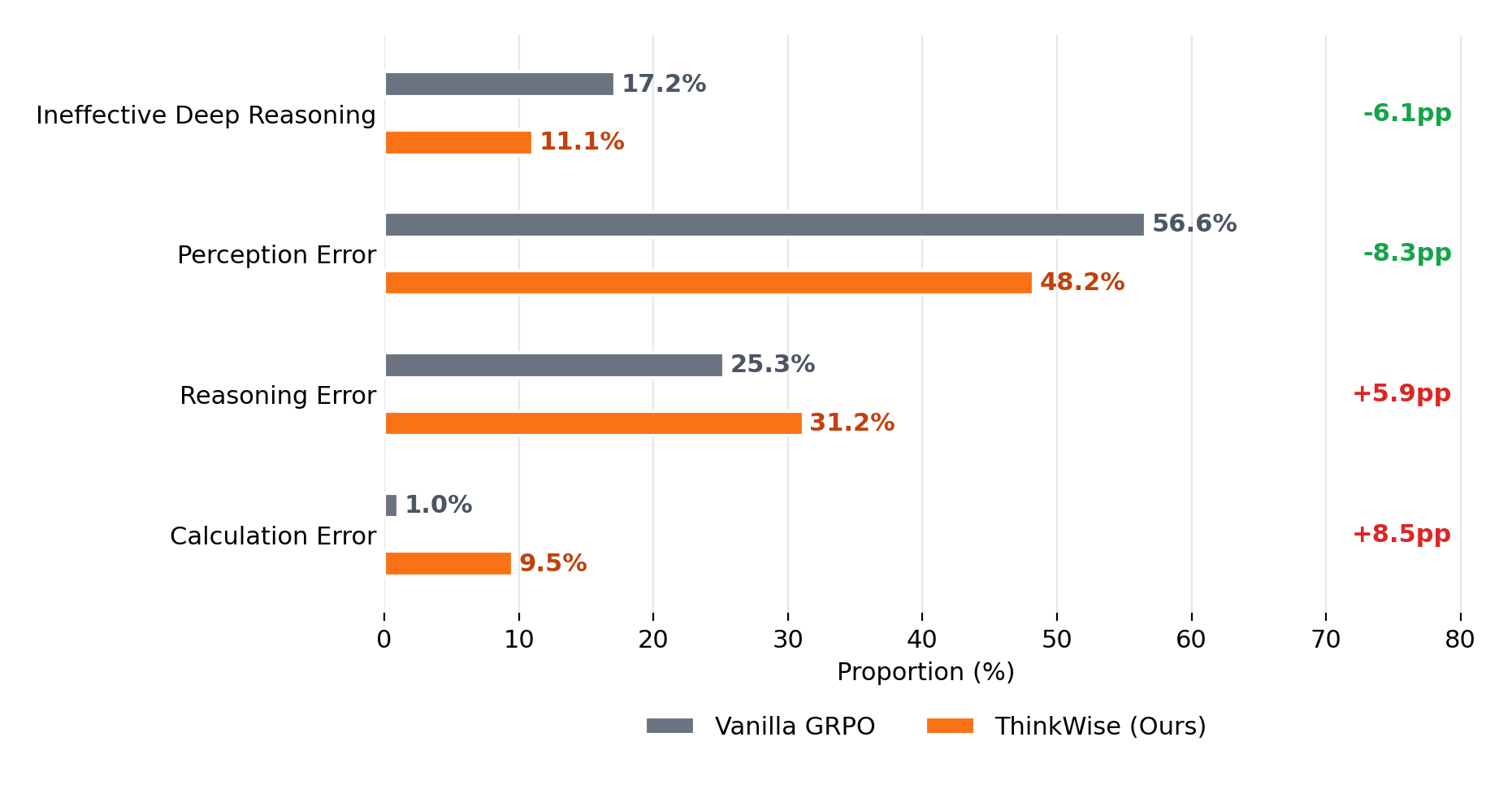}
    \caption{Annotated error category distributions. 
    }
    \label{fig:error_category}
\end{figure}
To better understand how ThinkWise affects multimodal reasoning behavior, we annotate model failures. Specifically, for each model, we select 200 error cases from Counting, MMK12, MMMU-Pro, and We-Math, and categorize them into four types: ineffective deep reasoning, perception error, reasoning error, and calculation error. \textbf{Ineffective Deep Reasoning} occurs when the reasoning process drifts away from visual evidence and becomes dominated by language priors or incorrect premises. \textbf{Perception Error} arises when the model mmisreads key visual information and builds subsequent reasoning on a wrong perceptual premise. \textbf{Reasoning Error} refers to cases where visual perception is correct, but the reasoning process is flawed. Finally, \textbf{Calculation Error} occurs when both perception and reasoning are correct, yet the final computation is incorrect. As shown in Figure~\ref{fig:error_category}, \textbf{ineffective deep reasoning} and \textbf{perception error} decrease by \textbf{6.1\%} and \textbf{8.4\%}, respectively, suggesting better use of visual evidence and less unproductive reasoning under weak grounding. By contrast, \textbf{reasoning error} and \textbf{calculation error} increase by \textbf{5.9\%} and \textbf{8.5\%}, respectively. 
This does not indicate weaker reasoning or numerical ability. Instead, once part of the perception-related failures is removed, the remaining errors become more concentrated in reasoning and computation. This suggests that ThinkWise shifts the primary bottleneck from visual grounding to how the model reasons and computes after identifying the relevant visual information.

\section{Related Work}

\subsection{Reinforcement Learning with Verifiable Rewards.}
RL-driven post-training has shown strong effectiveness for slow thinking in text-only large language models \cite{guo2025deepseek,shao2024deepseekmath,yu2025dapo,zhang2025step}, and is now being extended to multimodal reasoning. Recent works \cite{huang2025vision,peng2025lmm,meng2025mm,yang2025r1,liu2026towards} advance multimodal reasoning post-training through data construction, RL, and cross-modal modeling. These studies show that RL-driven post-training improves multimodal reasoning while motivating adaptive reasoning control. A related line of work studies adaptive control of reasoning depth. DAST \cite{shen2025dast} reduces redundant reasoning on simple problems through difficulty-adaptive rewards, while O1-Pruner \cite{luo2025o1} mitigates length disharmony in long-chain reasoning through fine-tuning. Similar ideas have also been extended to multimodal settings. FAST~\cite{xiao2025fast}  identifies overthinking in vision-language models and introduces adaptive thinking rewards to balance performance against token cost. These studies suggest that reasoning length can itself be treated as a controllable variable rather than a fixed byproduct of generation. 

\subsection{Large Multimodal Reasoning Models.}
Effective reasoning in large multimodal reasoning models depends not only on logical deduction, but also on whether the reasoning process is grounded in correct perceptual evidence. Prior work \cite{wang2025perception} shows that current LMRMs are largely inherited from text-only settings and typically optimize reasoning using supervision based solely on final-answer correctness. However, such supervision can conflate two fundamentally different sources of failure, namely reasoning failure and perception failure. Wang et al. \cite{wang2025perception} further show that a substantial portion of multimodal reasoning errors arise from perception-related failures rather than logical reasoning errors. Meanwhile, Reich et al. \cite{reich2023measuring} point out that answer correctness alone is insufficient to determine whether a prediction is genuinely grounded in visual evidence. Together, these findings suggest that final-answer supervision alone may be insufficient for modeling reasoning behavior in multimodal settings. Recent work has begun to explore internal model signals as a way to characterize grounding state. In LMRMs, grounding is closely related to internal attention behavior. iTaD \cite{xu2025mitigating} and subsequent studies \cite{Favero_2024_CVPR,ogasa2025hallucinatedspandetectionmultiview} report consistent correlations between attention-based features and grounding quality. At the same time, attention should not be interpreted as a direct measure of perceptual correctness, since attention weights alone do not fully explain Transformer behavior \cite{clark-etal-2019-bert,kobayashi-etal-2020-attention}. 

\section{Conclusion}

In this work, we propose \textit{ThinkWise}, a perception-aware RLVR framework that addresses reasoning allocation misalignment in large multimodal reasoning models. By incorporating trajectory-level perceptual signals via a lightweight endogenous proxy and introducing an instance-adaptive perception modeling mechanism, ThinkWise enables more accurate estimation of perceptual reliability. Building on this, we design a structured reward formulation that jointly models perception, correctness, and task difficulty to adaptively regulate reasoning depth. This transforms conventional outcome-only supervision into a perception-aware training paradigm, enabling more effective multimodal reasoning. Extensive experiments on multiple LMRM benchmarks demonstrate that ThinkWise consistently achieves significant improvements over strong RLVR baselines.

\bibliographystyle{ACM-Reference-Format}
\balance
\bibliography{refs}

@article{guo2025deepseek,
  title={Deepseek-r1: Incentivizing reasoning capability in llms via reinforcement learning},
  author={Guo, Daya and Yang, Dejian and Zhang, Haowei and Song, Junxiao and Wang, Peiyi and Zhu, Qihao and Xu, Runxin and Zhang, Ruoyu and Ma, Shirong and Bi, Xiao and others},
  journal={arXiv preprint arXiv:2501.12948},
  year={2025}
}

@article{shao2024deepseekmath,
  title={Deepseekmath: Pushing the limits of mathematical reasoning in open language models},
  author={Shao, Zhihong and Wang, Peiyi and Zhu, Qihao and Xu, Runxin and Song, Junxiao and Bi, Xiao and Zhang, Haowei and Zhang, Mingchuan and Li, YK and Wu, Yang and others},
  journal={arXiv preprint arXiv:2402.03300},
  year={2024}
}

@article{yu2025dapo,
  title={Dapo: An open-source llm reinforcement learning system at scale},
  author={Yu, Qiying and Zhang, Zheng and Zhu, Ruofei and Yuan, Yufeng and Zuo, Xiaochen and Yue, Yu and Dai, Weinan and Fan, Tiantian and Liu, Gaohong and Liu, Lingjun and others},
  journal={arXiv preprint arXiv:2503.14476},
  year={2025}
}

@article{huang2025vision,
  title={Vision-r1: Incentivizing reasoning capability in multimodal large language models},
  author={Huang, Wenxuan and Jia, Bohan and Zhai, Zijie and Cao, Shaosheng and Ye, Zheyu and Zhao, Fei and Xu, Zhe and Tang, Xu and Hu, Yao and Lin, Shaohui},
  journal={arXiv preprint arXiv:2503.06749},
  year={2025}
}

@article{peng2025lmm,
  title={Lmm-r1: Empowering 3b lmms with strong reasoning abilities through two-stage rule-based rl},
  author={Peng, Yingzhe and Zhang, Gongrui and Zhang, Miaosen and You, Zhiyuan and Liu, Jie and Zhu, Qipeng and Yang, Kai and Xu, Xingzhong and Geng, Xin and Yang, Xu},
  journal={arXiv preprint arXiv:2503.07536},
  year={2025}
}

@inproceedings{yang2025r1,
  title={R1-onevision: Advancing generalized multimodal reasoning through cross-modal formalization},
  author={Yang, Yi and He, Xiaoxuan and Pan, Hongkun and Jiang, Xiyan and Deng, Yan and Yang, Xingtao and Lu, Haoyu and Yin, Dacheng and Rao, Fengyun and Zhu, Minfeng and others},
  booktitle={Proceedings of the IEEE/CVF International Conference on Computer Vision},
  pages={2376--2385},
  year={2025}
}

@article{wang2025perception,
  title={Perception-aware policy optimization for multimodal reasoning},
  author={Wang, Zhenhailong and Guo, Xuehang and Stoica, Sofia and Xu, Haiyang and Wang, Hongru and Ha, Hyeonjeong and Chen, Xiusi and Chen, Yangyi and Yan, Ming and Huang, Fei and others},
  journal={arXiv preprint arXiv:2507.06448},
  year={2025}
}

@article{xiao2025fast,
  title={Fast-slow thinking grpo for large vision-language model reasoning},
  author={Xiao, Wenyi and Gan, Leilei},
  journal={arXiv preprint arXiv:2504.18458},
  year={2025}
}

@inproceedings{shen2025dast,
  title={Dast: Difficulty-adaptive slow-thinking for large reasoning models},
  author={Shen, Yi and Zhang, Jian and Huang, Jieyun and Shi, Shuming and Zhang, Wenjing and Yan, Jiangze and Wang, Ning and Wang, Kai and Liu, Zhaoxiang and Lian, Shiguo},
  booktitle={Proceedings of the 2025 Conference on Empirical Methods in Natural Language Processing: Industry Track},
  pages={2322--2331},
  year={2025}
}

@article{luo2025o1,
  title={O1-pruner: Length-harmonizing fine-tuning for o1-like reasoning pruning},
  author={Luo, Haotian and Shen, Li and He, Haiying and Wang, Yibo and Liu, Shiwei and Li, Wei and Tan, Naiqiang and Cao, Xiaochun and Tao, Dacheng},
  journal={arXiv preprint arXiv:2501.12570},
  year={2025}
}

@inproceedings{xu2025mitigating,
  title={Mitigating hallucinations in multi-modal large language models via image token attention-guided decoding},
  author={Xu, Xinhao and Chen, Hui and Lyu, Mengyao and Zhao, Sicheng and Xiong, Yizhe and Lin, Zijia and Han, Jungong and Ding, Guiguang},
  booktitle={Proceedings of the 2025 Conference of the Nations of the Americas Chapter of the Association for Computational Linguistics: Human Language Technologies (Volume 1: Long Papers)},
  pages={1571--1590},
  year={2025}
}

@inproceedings{reich2023measuring,
  title={Measuring faithful and plausible visual grounding in VQA},
  author={Reich, Daniel and Putze, Felix and Schultz, Tanja},
  booktitle={Findings of the Association for Computational Linguistics: EMNLP 2023},
  pages={3129--3144},
  year={2023}
}

@article{lu2023mathvista,
  title={Mathvista: Evaluating mathematical reasoning of foundation models in visual contexts},
  author={Lu, Pan and Bansal, Hritik and Xia, Tony and Liu, Jiacheng and Li, Chunyuan and Hajishirzi, Hannaneh and Cheng, Hao and Chang, Kai-Wei and Galley, Michel and Gao, Jianfeng},
  journal={arXiv preprint arXiv:2310.02255},
  year={2023}
}

@inproceedings{zhang2024mathverse,
  title={Mathverse: Does your multi-modal llm truly see the diagrams in visual math problems?},
  author={Zhang, Renrui and Jiang, Dongzhi and Zhang, Yichi and Lin, Haokun and Guo, Ziyu and Qiu, Pengshuo and Zhou, Aojun and Lu, Pan and Chang, Kai-Wei and Qiao, Yu and others},
  booktitle={European Conference on Computer Vision},
  pages={169--186},
  year={2024},
  organization={Springer}
}

@article{xiao2024logicvista,
  title={Logicvista: Multimodal llm logical reasoning benchmark in visual contexts},
  author={Xiao, Yijia and Sun, Edward and Liu, Tianyu and Wang, Wei},
  journal={arXiv preprint arXiv:2407.04973},
  year={2024}
}

@inproceedings{yue2025mmmu,
  title={Mmmu-pro: A more robust multi-discipline multimodal understanding benchmark},
  author={Yue, Xiang and Zheng, Tianyu and Ni, Yuansheng and Wang, Yubo and Zhang, Kai and Tong, Shengbang and Sun, Yuxuan and Yu, Botao and Zhang, Ge and Sun, Huan and others},
  booktitle={Proceedings of the 63rd Annual Meeting of the Association for Computational Linguistics (Volume 1: Long Papers)},
  pages={15134--15186},
  year={2025}
}

@article{deng2025boosting,
  title={Boosting the generalization and reasoning of vision language models with curriculum reinforcement learning},
  author={Deng, Huilin and Zou, Ding and Ma, Rui and Luo, Hongchen and Cao, Yang and Kang, Yu},
  journal={arXiv preprint arXiv:2503.07065},
  year={2025}
}

@article{chen2025advancing,
  title={Advancing multimodal reasoning: From optimized cold start to staged reinforcement learning},
  author={Chen, Shuang and Guo, Yue and Su, Zhaochen and Li, Yafu and Wu, Yulun and Chen, Jiacheng and Chen, Jiayu and Wang, Weijie and Qu, Xiaoye and Cheng, Yu},
  journal={arXiv preprint arXiv:2506.04207},
  year={2025}
}

@article{leng2025mmr1,
  title={Mmr1: Enhancing multimodal reasoning with variance-aware sampling and open resources},
  author={Leng, Sicong and Wang, Jing and Li, Jiaxi and Zhang, Hao and Hu, Zhiqiang and Zhang, Boqiang and Jiang, Yuming and Zhang, Hang and Li, Xin and Bing, Lidong and others},
  journal={arXiv preprint arXiv:2509.21268},
  year={2025}
}

@misc{chen2025r1v,
  author       = {Chen, Liang and Li, Lei and Zhao, Haozhe and Song, Yifan and Vinci},
  title        = {R1-V: Reinforcing Super Generalization Ability in Vision-Language Models with Less Than \$3},
  howpublished = {\url{https://github.com/Deep-Agent/R1-V}},
  note         = {Accessed: 2025-02-02},
  year         = {2025}
}

@article{shen2025vlm,
  title={Vlm-r1: A stable and generalizable r1-style large vision-language model},
  author={Shen, Haozhan and Liu, Peng and Li, Jingcheng and Fang, Chunxin and Ma, Yibo and Liao, Jiajia and Shen, Qiaoli and Zhang, Zilun and Zhao, Kangjia and Zhang, Qianqian and others},
  journal={arXiv preprint arXiv:2504.07615},
  year={2025}
}

@article{meng2025mm,
  title={Mm-eureka: Exploring the frontiers of multimodal reasoning with rule-based reinforcement learning},
  author={Meng, Fanqing and Du, Lingxiao and Liu, Zongkai and Zhou, Zhixiang and Lu, Quanfeng and Fu, Daocheng and Han, Tiancheng and Shi, Botian and Wang, Wenhai and He, Junjun and others},
  journal={arXiv preprint arXiv:2503.07365},
  year={2025}
}

@article{wang2025skywork,
  title={Skywork r1v2: Multimodal hybrid reinforcement learning for reasoning},
  author={Wang, Peiyu and Wei, Yichen and Peng, Yi and Wang, Xiaokun and Qiu, Weijie and Shen, Wei and Xie, Tianyidan and Pei, Jiangbo and Zhang, Jianhao and Hao, Yunzhuo and others},
  journal={arXiv preprint arXiv:2504.16656},
  year={2025}
}

@article{xiao2025perception,
  title={Perception-R1: Advancing Multimodal Reasoning Capabilities of MLLMs via Visual Perception Reward},
  author={Xiao, Tong and Xu, Xin and Huang, Zhenya and Gao, Hongyu and Liu, Quan and Liu, Qi and Chen, Enhong},
  journal={arXiv preprint arXiv:2506.07218},
  year={2025}
}

@article{zhu2025active,
  title={Active-o3: Empowering multimodal large language models with active perception via grpo},
  author={Zhu, Muzhi and Zhong, Hao and Zhao, Canyu and Du, Zongze and Huang, Zheng and Liu, Mingyu and Chen, Hao and Zou, Cheng and Chen, Jingdong and Yang, Ming and others},
  journal={arXiv preprint arXiv:2505.21457},
  year={2025}
}

@misc{xia2025visionaryr1mitigatingshortcutsvisual,
      title={Visionary-R1: Mitigating Shortcuts in Visual Reasoning with Reinforcement Learning}, 
      author={Jiaer Xia and Yuhang Zang and Peng Gao and Sharon Li and Kaiyang Zhou},
      year={2025},
      eprint={2505.14677},
      archivePrefix={arXiv},
      url={https://arxiv.org/abs/2505.14677}, 
}

@misc{wan2025srpoenhancingmultimodalllm,
      title={SRPO: Enhancing Multimodal LLM Reasoning via Reflection-Aware Reinforcement Learning}, 
      author={Zhongwei Wan and Zhihao Dou and Che Liu and Yu Zhang and Dongfei Cui and Qinjian Zhao and Hui Shen and Jing Xiong and Yi Xin and Yifan Jiang and Chaofan Tao and Yangfan He and Mi Zhang and Shen Yan},
      year={2025},
      eprint={2506.01713},
      archivePrefix={arXiv},
      url={https://arxiv.org/abs/2506.01713}, 
}

@InProceedings{Favero_2024_CVPR,
    author    = {Favero, Alessandro and Zancato, Luca and Trager, Matthew and Choudhary, Siddharth and Perera, Pramuditha and Achille, Alessandro and Swaminathan, Ashwin and Soatto, Stefano},
    title     = {Multi-Modal Hallucination Control by Visual Information Grounding},
    booktitle = {Proceedings of the IEEE/CVF Conference on Computer Vision and Pattern Recognition (CVPR)},
    month     = {June},
    year      = {2024},
    pages     = {14303-14312}
}

@misc{ogasa2025hallucinatedspandetectionmultiview,
      title={Hallucinated Span Detection with Multi-View Attention Features}, 
      author={Yuya Ogasa and Yuki Arase},
      year={2025},
      eprint={2504.04335},
      archivePrefix={arXiv},
      url={https://arxiv.org/abs/2504.04335}, 
}

@inproceedings{clark-etal-2019-bert,
    title = "What Does {BERT} Look at? An Analysis of {BERT}{'}s Attention",
    author = "Clark, Kevin  and
      Khandelwal, Urvashi  and
      Levy, Omer  and
      Manning, Christopher D.",
    editor = "Linzen, Tal  and
      Chrupa{\l}a, Grzegorz  and
      Belinkov, Yonatan  and
      Hupkes, Dieuwke",
    booktitle = "Proceedings of the 2019 ACL Workshop BlackboxNLP: Analyzing and Interpreting Neural Networks for NLP",
    month = aug,
    year = "2019",
    address = "Florence, Italy",
    publisher = "Association for Computational Linguistics",
    url = "https://aclanthology.org/W19-4828/",
    doi = "10.18653/v1/W19-4828",
    pages = "276--286"
}

@ARTICLE{9335490,
  author={Zhang, Peng-Fei and Li, Yang and Huang, Zi and Xu, Xin-Shun},
  journal={IEEE Transactions on Multimedia}, 
  title={Aggregation-Based Graph Convolutional Hashing for Unsupervised Cross-Modal Retrieval}, 
  year={2022},
  volume={24},
  number={},
  pages={466-479},
  doi={10.1109/TMM.2021.3053766}}

@inproceedings{kobayashi-etal-2020-attention,
    title = "Attention is Not Only a Weight: Analyzing Transformers with Vector Norms",
    author = "Kobayashi, Goro  and
      Kuribayashi, Tatsuki  and
      Yokoi, Sho  and
      Inui, Kentaro",
    editor = "Webber, Bonnie  and
      Cohn, Trevor  and
      He, Yulan  and
      Liu, Yang",
    booktitle = "Proceedings of the 2020 Conference on Empirical Methods in Natural Language Processing (EMNLP)",
    month = nov,
    year = "2020",
    address = "Online",
    publisher = "Association for Computational Linguistics",
    url = "https://aclanthology.org/2020.emnlp-main.574/",
    doi = "10.18653/v1/2020.emnlp-main.574",
    pages = "7057--7075"
}

@inproceedings{jian-etal-2025-look,
    title = "Look Again, Think Slowly: Enhancing Visual Reflection in Vision-Language Models",
    author = "Jian, Pu  and
      Wu, Junhong  and
      Sun, Wei  and
      Wang, Chen  and
      Ren, Shuo  and
      Zhang, Jiajun",
    editor = "Christodoulopoulos, Christos  and
      Chakraborty, Tanmoy  and
      Rose, Carolyn  and
      Peng, Violet",
    booktitle = "Proceedings of the 2025 Conference on Empirical Methods in Natural Language Processing",
    month = nov,
    year = "2025",
    address = "Suzhou, China",
    publisher = "Association for Computational Linguistics",
    url = "https://aclanthology.org/2025.emnlp-main.470/",
    doi = "10.18653/v1/2025.emnlp-main.470",
    pages = "9251--9270",
    ISBN = "979-8-89176-332-6"
}

@misc{chu2025qwenlookagainguiding,
      title={Qwen Look Again: Guiding Vision-Language Reasoning Models to Re-attention Visual Information}, 
      author={Xu Chu and Xinrong Chen and Guanyu Wang and Zhijie Tan and Kui Huang and Wenyu Lv and Tong Mo and Weiping Li},
      year={2025},
      eprint={2505.23558},
      archivePrefix={arXiv},
      url={https://arxiv.org/abs/2505.23558}, 
}

@ARTICLE{11644733,
  author={Zhang, Peng-Fei and Bai, Guangdong and Xu, Xin-Shun and Huang, Zi},
  journal={IEEE Transactions on Knowledge and Data Engineering}, 
  title={Data-Centric Challenges, Techniques, and Impacts: A Survey on Image Data Perturbation}, 
  year={2026},
  volume={38},
  number={9},
  pages={5736-5764},
  doi={10.1109/TKDE.2026.3709786}}

@inproceedings{Wu2026HowVB,
  title={How Vision Becomes Language: A Layer-wise Information-Theoretic Analysis of Multimodal Reasoning},
  author={Hongxuan Wu and Yukun Zhang and Xueqing Zhou},
  year={2026},
  url={https://api.semanticscholar.org/CorpusID:285659939}
}

@misc{chen2025sftrlearlyinvestigation,
      title={SFT or RL? An Early Investigation into Training R1-Like Reasoning Large Vision-Language Models}, 
      author={Hardy Chen and Haoqin Tu and Fali Wang and Hui Liu and Xianfeng Tang and Xinya Du and Yuyin Zhou and Cihang Xie},
      year={2025},
      eprint={2504.11468},
      archivePrefix={arXiv},
      url={https://arxiv.org/abs/2504.11468}, 
}

@misc{wei2025advancingmultimodalreasoningreinforcement,
      title={Advancing Multimodal Reasoning via Reinforcement Learning with Cold Start}, 
      author={Lai Wei and Yuting Li and Kaipeng Zheng and Chen Wang and Yue Wang and Linghe Kong and Lichao Sun and Weiran Huang},
      year={2025},
      eprint={2505.22334},
      archivePrefix={arXiv},
      url={https://arxiv.org/abs/2505.22334}, 
}

@misc{wang2024measuringmultimodalmathematicalreasoning,
      title={Measuring Multimodal Mathematical Reasoning with MATH-Vision Dataset}, 
      author={Ke Wang and Junting Pan and Weikang Shi and Zimu Lu and Mingjie Zhan and Hongsheng Li},
      year={2024},
      eprint={2402.14804},
      archivePrefix={arXiv},
      url={https://arxiv.org/abs/2402.14804}, 
}

@misc{qiao2024wemathdoeslargemultimodal,
      title={We-Math: Does Your Large Multimodal Model Achieve Human-like Mathematical Reasoning?}, 
      author={Runqi Qiao and Qiuna Tan and Guanting Dong and Minhui Wu and Chong Sun and Xiaoshuai Song and Zhuoma GongQue and Shanglin Lei and Zhe Wei and Miaoxuan Zhang and Runfeng Qiao and Yifan Zhang and Xiao Zong and Yida Xu and Muxi Diao and Zhimin Bao and Chen Li and Honggang Zhang},
      year={2024},
      eprint={2407.01284},
      archivePrefix={arXiv},
      url={https://arxiv.org/abs/2407.01284}, 
}

@misc{li2023superclevrvirtualbenchmarkdiagnose,
      title={Super-CLEVR: A Virtual Benchmark to Diagnose Domain Robustness in Visual Reasoning}, 
      author={Zhuowan Li and Xingrui Wang and Elias Stengel-Eskin and Adam Kortylewski and Wufei Ma and Benjamin Van Durme and Alan Yuille},
      year={2023},
      eprint={2212.00259},
      archivePrefix={arXiv},
      url={https://arxiv.org/abs/2212.00259}, 
}

@inproceedings{yao2025mmreason,
  title={Mmreason: An open-ended multi-modal multi-step reasoning benchmark for mllms toward agi},
  author={Yao, Huanjin and Huang, Jiaxing and Qiu, Yawen and Chen, Michael K and Liu, Wenzheng and Zhang, Wei and Zeng, Wenjie and Zhang, Xikun and Zhang, Jingyi and Song, Yuxin and others},
  booktitle={Proceedings of the IEEE/CVF International Conference on Computer Vision},
  pages={273--283},
  year={2025}
}

@article{li2025perception,
  title={Perception, reason, think, and plan: A survey on large multimodal reasoning models},
  author={Li, Yunxin and Liu, Zhenyu and Li, Zitao and Zhang, Xuanyu and Xu, Zhenran and Chen, Xinyu and Shi, Haoyuan and Jiang, Shenyuan and Wang, Xintong and Wang, Jifang and others},
  journal={arXiv preprint arXiv:2505.04921},
  year={2025}
}

@article{schulman2017proximal,
  title={Proximal policy optimization algorithms},
  author={Schulman, John and Wolski, Filip and Dhariwal, Prafulla and Radford, Alec and Klimov, Oleg},
  journal={arXiv preprint arXiv:1707.06347},
  year={2017}
}

@inproceedings{zhou2025proreason,
  title={PROREASON: Multi-Modal Proactive Reasoning with Decoupled Eyesight and Wisdom},
  author={Zhou, Jingqi and Wang, Sheng and Dong, Jingwei and Liu, Kai and Li, Lei and Gao, Jiahui and Jiang, Jiyue and Kong, Lingpeng and Wu, Chuan},
  booktitle={Proceedings of the 2025 Conference on Empirical Methods in Natural Language Processing},
  pages={31650--31679},
  year={2025}
}

@misc{hu2025reinforcestabilizingcriticfreepolicy,
      title={REINFORCE++: Stabilizing Critic-Free Policy Optimization with Global Advantage Normalization}, 
      author={Jian Hu and Jason Klein Liu and Haotian Xu and Wei Shen},
      year={2025},
      eprint={2501.03262},
      archivePrefix={arXiv},
      url={https://arxiv.org/abs/2501.03262}, 
}

@article{zhang2025step,
  title={A step toward world models: A survey on robotic manipulation},
  author={Zhang, Peng-Fei and Cheng, Ying and Sun, Xiaofan and Wang, Shijie and Li, Fengling and Zhu, Lei and Shen, Heng Tao},
  journal={arXiv preprint arXiv:2511.02097},
  year={2025}
}

@inproceedings{zhang2024universal,
  title={Universal adversarial perturbations for vision-language pre-trained models},
  author={Zhang, Peng-Fei and Huang, Zi and Bai, Guangdong},
  booktitle={Proceedings of the 47th International ACM SIGIR Conference on Research and Development in Information Retrieval},
  pages={862--871},
  year={2024}
}

@inproceedings{liu2026towards,
  title={Towards Practical LLM Unlearning: Efficient, Modular, and Retain-Free},
  author={Liu, Peng and Zhang, Peng-Fei and Qu, Jianfeng and Li, Ximing and Li, Zhixu and Zhao, Pengpeng},
  booktitle={Proceedings of the ACM Web Conference 2026},
  pages={2846--2857},
  year={2026}
}

@article{zhang2026hierarchical,
  title={Hierarchical Refinement of Universal Multimodal Attacks on Vision-Language Models},
  author={Zhang, Peng-Fei and Bai, Guangdong and Huang, Zi},
  journal={IEEE Transactions on Multimedia},
  year={2026},
  publisher={IEEE}
}

\clearpage

\end{document}